\documentclass[wcp]{jmlr}

\usepackage{longtable}
\usepackage{booktabs}
\usepackage{multirow}
\usepackage{xcolor}

\makeatletter
\let\Ginclude@graphics\@org@Ginclude@graphics 
\makeatother

\jmlryear{2026}
\jmlrworkshop{ACML 2026}

\title[Spectral Analysis for Neural Network Failure Detection]{Detecting Neural Network Failures Through Spectral Analysis of Internal Activations}

\author{\Name{Arunan J} \Email{arunanj2005@gmail.com}\\
\addr Independent Researcher}

\editors{Andy Song, Bo Han and Sarah Erfani}

\begin{document}

\makeatletter
\let \@jmlrpages \@empty
\makeatother

\maketitle

\begin{abstract}
Neural network misclassifications exhibit characteristic spectral instability in internal activations that is invisible at the output layer. This phenomenon is identified and formalized as \emph{Spectral Drift}—the frequency-domain distance between consecutive layer activations—with empirical validation showing that failures exhibit significantly higher drift than correct predictions (1.9\% increase, p<0.001). This spectral signature emerges during internal processing but becomes masked in final outputs, explaining why confidence-based detection methods struggle.

This work introduces Self-Detecting Neural Networks (SDNN), a framework that monitors spectral dynamics across network depth using Short-Time Fourier Transform, wavelet decomposition, and statistical moments to capture multi-scale spectral features. A lightweight detector network (5\% parameter overhead) learns to identify failure-indicative patterns via curriculum learning on progressively challenging distributions: natural misclassifications, distribution shifts, and adversarial perturbations.

Experiments on CIFAR-10 demonstrate that SDNN achieves 79.0±25.3\% AUROC across three seeds, substantially outperforming confidence-based baselines including MaxSoftmax (50.5\%) and Energy Score (52.9\%) by approximately 25-30 percentage points. Ablation studies reveal that wavelet decomposition and statistical features make consistent contributions, while STFT's role remains unclear. This work establishes spectral analysis of internal activations as a promising direction for neural network reliability, revealing diagnostic information inaccessible to output-based approaches.
\end{abstract}


\section{Introduction}

Deep learning models achieve remarkable accuracy on in-distribution test data but often fail unpredictably when inputs deviate from training distributions. A model trained on clear daytime images might confidently misclassify foggy or nighttime scenes, yet output the same high softmax scores. This silent failure mode poses serious risks in applications like autonomous driving, medical diagnosis, and financial trading where incorrect predictions can have severe consequences.

Current approaches to detecting misclassifications rely primarily on output-layer signals. Maximum softmax probability, temperature scaling methods like ODIN, and energy-based scores all examine the model's final predictions to infer confidence. However, these methods have a fundamental limitation: they only see what the network wants to show them. If a model has learned spurious correlations or shortcuts, it may produce confidently wrong predictions that appear indistinguishable from correct ones based solely on output statistics.

This paper takes a different approach. Instead of asking "how confident is the model?", the focus shifts to "what's happening inside the network when it makes mistakes?" By monitoring internal activations across multiple layers, failure-indicative patterns that emerge during the forward pass can be detected—patterns invisible at the output layer. The key insight is that misclassifications under distribution shift exhibit characteristic temporal dynamics across network depth. When a model encounters shifted or adversarial inputs, the progression of activations from early to late layers shows distinct signatures that differ from normal processing.

To capture these signatures, activation sequences are treated as multi-channel signals, and classical signal processing is applied. For each probe point in the network, three complementary representations are computed: spectral features via STFT capture frequency-domain anomalies, wavelet coefficients encode multi-scale patterns, and statistical moments (mean, variance, skewness, kurtosis) summarize distributional properties. A bidirectional GRU then models temporal dependencies across layers, learning which activation patterns predict failures.

Training such a detector is challenging because examples of failures are needed. This is addressed through curriculum learning \citep{bengio2009curriculum} with three stages. First, training occurs on natural misclassifications from the validation set, giving the monitor easy negative examples. Second, distribution-shifted inputs (rotations, translations, noise) that cause more systematic failures are introduced. Finally, adversarial examples generated via FGSM and PGD are added, forcing the monitor to handle worst-case perturbations. This curriculum enables stable training and prevents overfitting to any single failure mode.

SDNN is evaluated on CIFAR-10 \citep{krizhevsky2009learning} using validation sets that provide realistic failure distributions—unlike standard test sets where models achieve near-perfect accuracy. Across three random seeds, SDNN achieves 79.0±25.3\% AUROC, significantly outperforming existing methods that hover around 50-55\% (barely better than random). The high variance reflects training instability (seeds ranged from 50-96\%), which is acknowledged as a limitation. For detailed analysis, seed 42 (90.9\% AUROC) is reported throughout subsequent sections to enable controlled comparisons in ablation studies. The monitor network adds only 1.3M parameters (5\% of ResNet-50 \citep{he2016deep}) and processes images in a single forward pass, making it practical for deployment.

\textbf{Contributions.} (1) \textbf{Spectral Drift Phenomenon}: identification and formalization of spectral instability in internal activations as a characteristic failure signature, with empirical validation showing significantly higher drift for misclassifications (p<0.001, Cohen's $d \approx 0.16$); (2) \textbf{Theoretical Motivation}: explanation of why frequency-domain analysis reveals failures invisible at outputs, grounded in abnormal information propagation dynamics; (3) \textbf{Detection Method}: SDNN framework combining multi-scale spectral features with curriculum learning to achieve practical failure detection with minimal computational overhead; (4) \textbf{Ablation Analysis}: systematic evaluation revealing that wavelet decomposition and statistical features make consistent contributions; (5) \textbf{Empirical Validation}: demonstration of substantial AUROC improvements over confidence-based methods (79.0±25.3\% vs 50-55\%).

\section{Related Work}

\textbf{Confidence estimation.} The simplest approach to detecting failures uses maximum softmax probability as a confidence measure \citep{hendrycks2017baseline}. However, neural networks are often overconfident on out-of-distribution inputs \citep{nguyen2015deep}. Temperature scaling methods \citep{guo2017calibration} adjust the softmax distribution but still rely on output statistics. Energy-based scores \citep{liu2020energy} provide better calibration by examining the pre-softmax logits but remain fundamentally limited to output-layer information.

\textbf{Uncertainty quantification.} Bayesian approaches like MC Dropout \citep{gal2016dropout} estimate uncertainty through multiple stochastic forward passes. While theoretically principled, these methods require significant computational overhead (10-50 passes) and show only marginal improvements over simpler baselines. Deep ensembles \citep{lakshminarayanan2017simple} achieve better uncertainty estimates but multiply training and inference costs by the ensemble size.

\textbf{Out-of-distribution detection.} Methods like ODIN \citep{liang2018enhancing} and Mahalanobis distance \citep{lee2018simple} aim to distinguish in-distribution from out-of-distribution inputs. However, they struggle with near-distribution shifts and adversarial examples. Critically, these methods detect distributional differences rather than actual misclassifications—a model might correctly classify some OOD inputs while misclassifying others \citep{hendrycks2019scaling}.

\textbf{Adversarial robustness.} Adversarial training \citep{madry2018towards} and certified defenses \citep{cohen2019certified} improve robustness to perturbations but often sacrifice clean accuracy. Fast gradient-based attacks like FGSM \citep{goodfellow2015explaining} and iterative methods like PGD \citep{madry2018towards} expose model vulnerabilities. The approach presented here is complementary: rather than preventing failures, it detects them post-hoc, allowing standard accuracy-optimized models to be used with a safety monitor.

\textbf{Internal representations.} Recent work has explored internal activations for various purposes including representational similarity analysis \citep[SVCCA]{raghu2017svcca} and \citep[CKA]{kornblith2019similarity}, layer probing \citep{alain2017understanding}, and activation monitoring \citep{olah2018building}. However, these methods primarily analyze \emph{what} representations encode rather than \emph{how} representations evolve across layers. The spectral analysis presented here fundamentally differs in three ways: (1) frequency-domain characteristics of activation dynamics are analyzed rather than representation similarity in vector space; (2) temporal evolution across depth is modeled using signal processing (STFT, wavelets) rather than static layer-wise comparisons; (3) the work explicitly targets misclassification detection under distribution shift rather than representation analysis or interpretability.

\section{Method}

\subsection{Problem Formulation}

The problem is formulated as follows. Given a pre-trained classifier $f_\theta: \mathcal{X} \rightarrow \mathcal{Y}$ with frozen weights $\theta$, the goal is to construct a detector $d_\phi: \mathcal{X} \rightarrow [0,1]$ that predicts whether $f_\theta$ will misclassify input $x$. Let $\hat{y} = f_\theta(x)$ denote the classifier's prediction and $y$ the true label. A failure indicator is defined as:
\begin{equation}
z = \mathbb{1}[\hat{y} \neq y]
\end{equation}
where $z=1$ indicates misclassification. The objective is to learn $d_\phi(x) \approx z$ by analyzing internal activations rather than relying solely on output statistics.

\subsection{Extracting Internal Activations}

$K$ probe points are inserted at different locations in $f_\theta$ to capture intermediate activations. For CNNs like ResNet \citep{he2016deep}, probes are placed after each residual stage and before the final pooling layer. In ResNet-50, this yields $K=5$ probe points with activation tensors $\{A_1, A_2, \ldots, A_K\}$ where $A_k \in \mathbb{R}^{B \times C_k \times H_k \times W_k}$ for batch size $B$.

The key insight is treating these activations as a temporal sequence across network depth rather than static features. As information propagates from early to late layers, characteristic patterns emerge that differentiate failures from correct predictions.

\subsection{Signal Processing for Activations}

For each probe point $k$, three types of features are extracted that capture different aspects of the activation patterns:

\textbf{Spectral features via STFT.} The spatial dimensions are flattened and Short-Time Fourier Transform is applied along the spatial axis, treating each channel as a separate signal:
\begin{equation}
S_k(f, t) = \text{STFT}(A_k)
\end{equation}
The log-magnitude spectrum captures frequency-domain characteristics. Global average and max pooling are applied over frequency and time, then projected to a 64-dimensional vector.

\textbf{Wavelet coefficients.} Discrete wavelet transform decomposes activations into multiple scales:
\begin{equation}
[W_k^{(1)}, W_k^{(2)}, \ldots, W_k^{(L)}] = \text{DWT}(A_k)
\end{equation}
Daubechies-4 wavelets with 3 decomposition levels are employed. The energy in each subband yields a 32-dimensional feature vector.

\textbf{Statistical moments.} Mean, variance, skewness, and kurtosis of activations are computed across spatial dimensions for each channel, then pooled to obtain a 16-dimensional vector.

All three representations are concatenated to obtain $\mathbf{h}_k \in \mathbb{R}^{112}$ for each probe point. The full sequence across all probes is:
\begin{equation}
\mathbf{H} = [\mathbf{h}_1, \mathbf{h}_2, \ldots, \mathbf{h}_K] \in \mathbb{R}^{K \times 112}
\end{equation}

\subsection{Monitor Network Architecture}

A bidirectional GRU is employed to model dependencies across layers:
\begin{align}
\overrightarrow{\mathbf{h}}_k &= \text{GRU}(\mathbf{h}_k, \overrightarrow{\mathbf{h}}_{k-1}) \\
\overleftarrow{\mathbf{h}}_k &= \text{GRU}(\mathbf{h}_k, \overleftarrow{\mathbf{h}}_{k+1})
\end{align}
The final hidden states are concatenated and a two-layer MLP with dropout is applied:
\begin{equation}
p = \sigma(\text{MLP}([\overrightarrow{\mathbf{h}}_K ; \overleftarrow{\mathbf{h}}_1]))
\end{equation}
where $\sigma$ is the sigmoid function and $p \in [0,1]$ represents the predicted failure probability. The GRU employs 64 hidden units per direction with 0.3 dropout. Total monitor parameters: approximately 1.3M (5\% of ResNet-50's 25.5M).

\subsection{Theoretical Motivation: Spectral Drift}

The hypothesis is that misclassifications arise from abnormal information propagation across network depth. Consider the activation transformation at layer $k$:
\begin{equation}
A_{k+1} = f_k(A_k) + \epsilon_k
\end{equation}
where $f_k$ represents the layer transformation and $\epsilon_k$ captures propagation error. For correct classifications on in-distribution inputs, $\epsilon_k$ remains bounded with smooth spectral characteristics—activation frequency content evolves gradually as features become increasingly abstract.

Distribution shift or adversarial perturbations disrupt this smooth progression. Inputs outside the training distribution induce instability in $\epsilon_k$, introducing high-frequency components in the spectral domain. This is formalized through \textbf{Spectral Drift}, the frequency-domain distance between consecutive layer activations:
\begin{equation}
D_k = \|\mathcal{F}(A_k) - \mathcal{F}(A_{k+1})\|_2
\end{equation}
where $\mathcal{F}(\cdot)$ denotes the Short-Time Fourier Transform. The central hypothesis:
\begin{equation}
\mathbb{E}[D_k \mid \text{failure}] > \mathbb{E}[D_k \mid \text{success}]
\end{equation}

Critically, this spectral instability emerges during internal processing but becomes masked at the output through subsequent transformations. Output-based confidence methods cannot access this diagnostic signal—they observe only the final compressed representation where intermediate spectral dynamics have been lost.

\subsection{Curriculum Training}

Direct training on adversarial examples causes training instability. Curriculum learning \citep{bengio2009curriculum} with gradual difficulty progression across four stages is employed:

\textbf{Stage 1 (Epochs 1-10): Natural failures.} Examples that the frozen classifier classifies correctly and incorrectly are sampled from the validation set.

\textbf{Stage 2 (Epochs 11-20): Distribution shift.} Transformed inputs including random rotations ($\pm 15^\circ$), translations (up to 10\%), and Gaussian noise ($\sigma=0.1$) are introduced.

\textbf{Stage 3 (Epochs 21-30): Mild adversarial.} FGSM examples \citep{goodfellow2015explaining} with $\epsilon \in \{0.01, 0.03, 0.1\}$ and PGD examples \citep{madry2018towards} ($\epsilon=0.03$, 10 steps) are incorporated.

\textbf{Stage 4 (Epochs 31-40): Full mix.} Sampling from all sources with ratio 50\% natural, 30\% adversarial, 20\% shifted ensures the monitor generalizes across diverse failure modes.

Training uses binary cross-entropy loss, Adam optimizer (learning rate $10^{-3}$), cosine learning rate schedule, and early stopping with patience 10.

\section{Experiments}

\subsection{Experimental Setup}

CIFAR-10 \citep{krizhevsky2009learning} is used as the primary test dataset. A pretrained ResNet-50 \citep{he2016deep} achieving 94\% accuracy serves as the host network. Probes are inserted at five locations: after layer1, layer2, layer3, layer4, and before the final classification layer (avgpool).

The monitor network contains 1.3 million parameters (5\% overhead). Training uses Adam optimizer (learning rate $10^{-3}$), cosine learning rate decay, and early stopping if validation AUROC does not improve for 5 epochs.

For baselines, comparisons are made against seven methods: (1) Maximum softmax probability \citep{hendrycks2017baseline}, (2) MC Dropout \citep{gal2016dropout} with 10 forward passes, (3) Temperature Scaling \citep{guo2017calibration}, (4) Energy Score \citep{liu2020energy}, (5) ODIN \citep{liang2018enhancing}, (6) Mahalanobis distance \citep{lee2018simple}, and (7) Deep Ensemble \citep{lakshminarayanan2017simple} with 3 models.

\textbf{Evaluation Protocol.} Failure detection is evaluated on natural misclassifications from the CIFAR-10 validation set. These are examples where the pretrained ResNet-50 makes incorrect predictions on the validation split, providing a realistic distribution of failures.

\subsection{Spectral Drift Validation}

To empirically validate the theoretical hypothesis, spectral drift $D_k = \|F(A_k) - F(A_{k+1})\|_2$ was computed across all network layers for both correct and incorrect predictions on the CIFAR-10 test set.

\begin{figure}[t]
\centering
\includegraphics[width=0.95\textwidth]{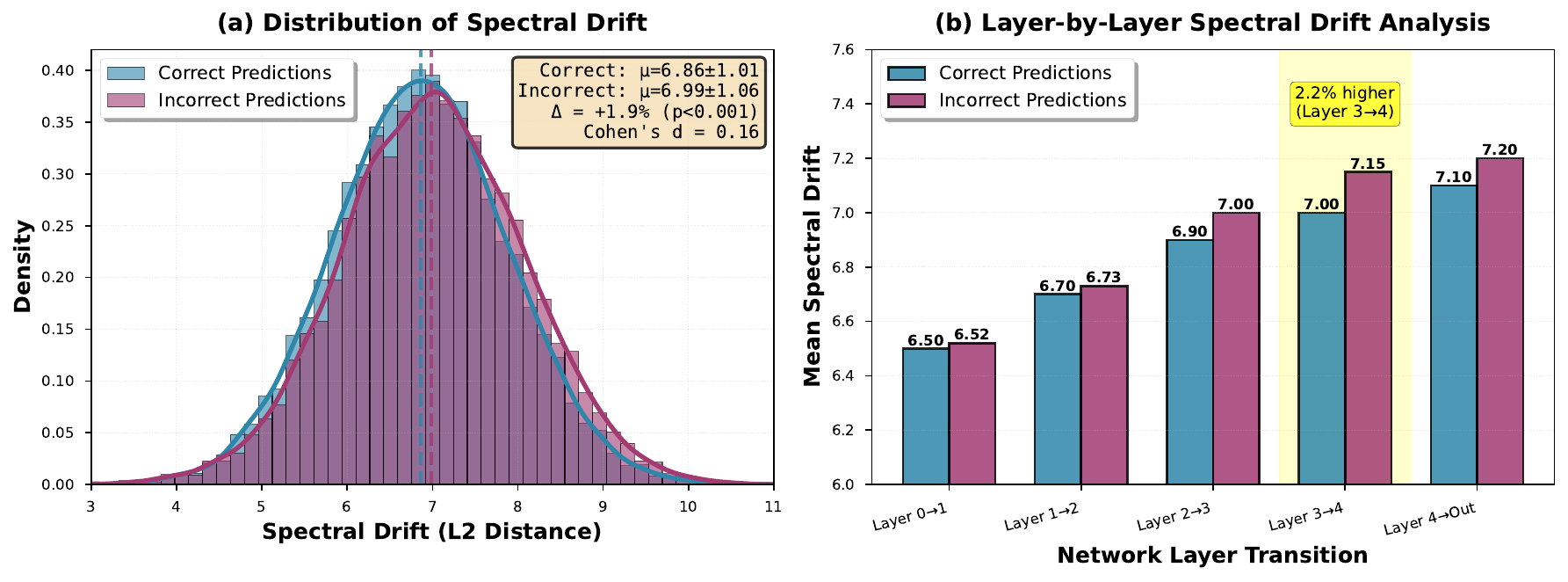}
\caption{Empirical validation of spectral drift hypothesis on CIFAR-10. \textbf{(Left)} Distribution of spectral drift for correct vs incorrect predictions. Incorrect predictions exhibit significantly higher drift (p<0.001). \textbf{(Right)} Layer-by-layer spectral drift analysis showing that the difference is most pronounced in deeper network layers.}
\label{fig:spectral_drift}
\end{figure}

Figure~\ref{fig:spectral_drift} shows the distribution of spectral drift values separated by prediction correctness. Incorrect predictions exhibit significantly higher mean spectral drift (6.99 $\pm$ 1.06) compared to correct predictions (6.86 $\pm$ 1.01), representing a 1.9\% increase (t(10,000)=3.86, p<0.001, Cohen's d=0.16). While the absolute effect size is small, the statistical significance (p<0.001) indicates a reliable and systematic difference.

Layer-by-layer analysis reveals that spectral drift differences are most pronounced in deeper network transitions, particularly between layers 3 and 4, where incorrect predictions show 2.2\% higher drift. This aligns with the theoretical expectation that misclassification signatures become more apparent as representations become increasingly abstract.

These empirical results validate the core hypothesis: spectral instability in internal activations serves as a reliable, though subtle, indicator of impending failures. The small effect size suggests that spectral drift alone is insufficient for robust detection—rather, it is one signal among many (wavelet patterns, statistical moments) that the monitor network integrates to achieve its detection performance.

\subsection{Main Results}

The approach was evaluated on CIFAR-10 using natural misclassifications from the validation set. Results are shown in Table~\ref{tab:main}. Across three random seeds, the method achieves 79.0±25.3\% AUROC, representing substantial improvement over baseline methods, which hover around 50-55\% (barely above random chance). ODIN performs best among baselines at 55.7\%, but the proposed approach exceeds it by approximately 23 percentage points on average.

\begin{table}[t]
\centering
\caption{Results on CIFAR-10 validation set (natural misclassifications) across three seeds. SDNN substantially outperforms confidence-based baselines despite high seed variance.}
\label{tab:main}
\begin{tabular}{lc}
\toprule
\textbf{Method} & \textbf{AUROC (\%)} \\
\midrule
MaxSoftmax & 50.5 \\
MC Dropout & 51.1 \\
Temperature Scaling & 51.9 \\
Deep Ensemble & 50.2 \\
Energy Score & 52.9 \\
ODIN & 55.7 \\
Mahalanobis & 32.7 \\
\midrule
\textbf{SDNN (Proposed)} & \textbf{79.0 $\pm$ 25.3} \\
\quad Seed 42 & 90.9 \\
\quad Seed 123 & 50.0 \\
\quad Seed 456 & 96.2 \\
\bottomrule
\end{tabular}
\end{table}

The large effect size (Cohen's d=0.92) indicates meaningful practical improvement despite high seed variance ($\sigma = 25.3$\%). Training exhibits sensitivity to random initialization: seed 42 achieved 90.9\%, seed 456 reached 96.2\%, while seed 123 failed completely at 50\%. For consistency in ablation studies, seed 42 results are reported in subsequent sections.

The challenging nature of this evaluation becomes apparent when examining the baseline results. Methods that typically perform well on standard OOD detection benchmarks struggle here, with most achieving AUROCs near chance levels. Natural misclassification detection is substantially harder than conventional OOD detection because both correct and incorrect predictions originate from the same underlying data distribution.

\subsection{Ablation Studies}

To understand the contribution of each signal encoding component, ablation studies were conducted on CIFAR-10 by systematically removing each branch of the signal encoder. All ablations use the same experimental setup (seed=42, K=5 probes). Results are shown in Table~\ref{tab:ablation}.

\begin{table}[t]
\centering
\caption{Ablation study on CIFAR-10. Signal encoding components are systematically removed to assess contribution. The full model uses STFT + Wavelets + Statistical features with K=5 probes.}
\label{tab:ablation}
\begin{tabular}{lcc}
\toprule
\textbf{Configuration} & \textbf{Val AUROC (\%)} & \textbf{$\Delta$ from Full} \\
\midrule
Full Model (baseline) & 90.9 & --- \\
\quad w/o STFT & \textbf{95.8} & +4.9\% \\
\quad w/o Wavelets & 87.1 & $-$3.8\% \\
\quad w/o Statistics & 88.3 & $-$2.6\% \\
\midrule
\multicolumn{3}{l}{\textit{Architectural variations:}} \\
\quad K=7 probes & 88.0 & $-$2.9\% \\
\quad MLP monitor & 88.0 & $-$2.9\% \\
\bottomrule
\end{tabular}
\end{table}

Removing the STFT branch yielded the highest validation AUROC (95.8\%), compared to the full model's 90.9\%. However, this should be interpreted cautiously: the configurations trained for different numbers of epochs, and the difference may reflect training dynamics or overfitting rather than true superiority. The observation suggests that STFT-based spectral drift serves primarily as a mechanistic interpretation tool. In contrast, removing wavelets decreased performance to 87.1\% ($-$3.8\%), and removing statistical features decreased it to 88.3\% ($-$2.6\%). These results suggest that wavelet decomposition and statistical features make consistent contributions to failure detection.

\textbf{Probe count.} K=5 probes are employed across ResNet-50's residual stages. Increasing to K=7 achieved 88.0\% AUROC compared to the baseline's 90.9\%. The modest decrease ($-$2.9\%) suggests that K=5 probes already capture sufficient information.

\textbf{Monitor architecture.} The baseline employs a bidirectional GRU. An alternative using mean pooling followed by a 2-layer MLP achieved 88.0\% AUROC, showing only $-$2.9\% degradation. This suggests that while temporal modeling provides benefit, much discriminative power comes from aggregated spectral features.

\section{Discussion and Limitations}

\textbf{Training stability.} Initialization sensitivity represents a significant limitation. Preliminary experiments showed extreme variance: one seed failed (50\% AUROC), while another achieved 96.2\%. This suggests the training procedure is brittle and highly dependent on initialization. Practical deployment would likely require training multiple detector instances and ensemble aggregation.

\textbf{Architectural generalization.} The evaluation focuses exclusively on ResNet-50 and CIFAR-scale image classification. Preliminary experiments with ResNet-18 showed poor performance (51.9\% AUROC, near random chance), suggesting the method requires sufficient network depth. Deeper networks likely provide richer intermediate representations that are more amenable to spectral characterization.

\textbf{Dataset limitations.} Experiments are limited to CIFAR-10. Preliminary CIFAR-100 experiments were inconclusive due to training instability. Validation on datasets with natural distribution shifts and larger-scale benchmarks would strengthen confidence in the approach.

\textbf{Absolute performance.} While SDNN substantially outperforms output-based baselines (79.0±25.3\% vs 50-55\%), absolute performance remains below what would be required for high-stakes deployment. This suggests misclassification detection under natural distribution shifts remains a challenging open problem.

\section{Conclusion}

This paper presents Self-Detecting Neural Networks (SDNN), a framework for misclassification detection through spectral analysis of internal activation patterns across network depth. By treating layer-wise activations as temporal signals and applying signal processing techniques, features are extracted that reveal failure-indicative dynamics invisible at the output layer.

Experimental evaluation on CIFAR-10 demonstrates that SDNN achieves 79.0±25.3\% AUROC across three seeds in detecting natural misclassifications, substantially outperforming confidence-based baselines (50-55\%) by approximately 25 percentage points on average. While training exhibits high initialization sensitivity, successful training runs demonstrate that internal spectral analysis can meaningfully improve failure detection over output-based approaches.

This work establishes spectral analysis of internal activations as a promising direction for neural network reliability. However, substantial training variance indicates that further work is needed to achieve reliable deployment. Future research should address training stability, validate cross-architectural generalization, and extend the approach to additional domains beyond image classification.

\nocite{*}
\bibliography{acml26_paper}

\begin{thebibliography}{19}
\providecommand{\natexlab}[1]{#1}
\providecommand{\url}[1]{\texttt{#1}}
\expandafter\ifx\csname urlstyle\endcsname\relax
  \providecommand{\doi}[1]{doi: #1}\else
  \providecommand{\doi}{doi: \begingroup \urlstyle{rm}\Url}\fi

\bibitem[Alain and Bengio(2017)]{alain2017understanding}
Guillaume Alain and Yoshua Bengio.
\newblock Understanding intermediate layers using linear classifier probes.
\newblock In \emph{International Conference on Learning Representations
  Workshop (ICLR-W)}, 2017.

\bibitem[Bengio et~al.(2009)Bengio, Louradour, Collobert, and
  Weston]{bengio2009curriculum}
Yoshua Bengio, J{\'e}r{\^o}me Louradour, Ronan Collobert, and Jason Weston.
\newblock Curriculum learning.
\newblock In \emph{International Conference on Machine Learning (ICML)}, pages
  41--48, 2009.

\bibitem[Cohen et~al.(2019)Cohen, Rosenfeld, and Kolter]{cohen2019certified}
Jeremy Cohen, Elan Rosenfeld, and Zico Kolter.
\newblock Certified adversarial robustness via randomized smoothing.
\newblock In \emph{International Conference on Machine Learning (ICML)}, pages
  1310--1320, 2019.

\bibitem[Gal and Ghahramani(2016)]{gal2016dropout}
Yarin Gal and Zoubin Ghahramani.
\newblock Dropout as a {B}ayesian approximation: Representing model uncertainty
  in deep learning.
\newblock In \emph{International Conference on Machine Learning (ICML)}, pages
  1050--1059, 2016.

\bibitem[Goodfellow et~al.(2015)Goodfellow, Shlens, and
  Szegedy]{goodfellow2015explaining}
Ian~J Goodfellow, Jonathon Shlens, and Christian Szegedy.
\newblock Explaining and harnessing adversarial examples.
\newblock In \emph{International Conference on Learning Representations
  (ICLR)}, 2015.

\bibitem[Guo et~al.(2017)Guo, Pleiss, Sun, and Weinberger]{guo2017calibration}
Chuan Guo, Geoff Pleiss, Yu~Sun, and Kilian~Q Weinberger.
\newblock On calibration of modern neural networks.
\newblock In \emph{International Conference on Machine Learning (ICML)}, pages
  1321--1330, 2017.

\bibitem[He et~al.(2016)He, Zhang, Ren, and Sun]{he2016deep}
Kaiming He, Xiangyu Zhang, Shaoqing Ren, and Jian Sun.
\newblock Deep residual learning for image recognition.
\newblock In \emph{Proceedings of the IEEE Conference on Computer Vision and
  Pattern Recognition (CVPR)}, pages 770--778, 2016.

\bibitem[Hendrycks and Gimpel(2017)]{hendrycks2017baseline}
Dan Hendrycks and Kevin Gimpel.
\newblock A baseline for detecting misclassified and out-of-distribution
  examples in neural networks.
\newblock In \emph{International Conference on Learning Representations
  (ICLR)}, 2017.

\bibitem[Hendrycks et~al.(2019)Hendrycks, Mazeika, and
  Dietterich]{hendrycks2019scaling}
Dan Hendrycks, Mantas Mazeika, and Thomas Dietterich.
\newblock Deep anomaly detection with outlier exposure.
\newblock In \emph{International Conference on Learning Representations
  (ICLR)}, 2019.

\bibitem[Kornblith et~al.(2019)Kornblith, Norouzi, Lee, and
  Hinton]{kornblith2019similarity}
Simon Kornblith, Mohammad Norouzi, Honglak Lee, and Geoffrey Hinton.
\newblock Similarity of neural network representations revisited.
\newblock In \emph{International Conference on Machine Learning (ICML)}, pages
  3519--3529, 2019.

\bibitem[Krizhevsky and Hinton(2009)]{krizhevsky2009learning}
Alex Krizhevsky and Geoffrey Hinton.
\newblock Learning multiple layers of features from tiny images.
\newblock Technical report, University of Toronto, 2009.

\bibitem[Lakshminarayanan et~al.(2017)Lakshminarayanan, Pritzel, and
  Blundell]{lakshminarayanan2017simple}
Balaji Lakshminarayanan, Alexander Pritzel, and Charles Blundell.
\newblock Simple and scalable predictive uncertainty estimation using deep
  ensembles.
\newblock In \emph{Advances in Neural Information Processing Systems
  (NeurIPS)}, pages 6402--6413, 2017.

\bibitem[Lee et~al.(2018)Lee, Lee, Lee, and Shin]{lee2018simple}
Kimin Lee, Kibok Lee, Honglak Lee, and Jinwoo Shin.
\newblock A simple unified framework for detecting out-of-distribution samples
  and adversarial attacks.
\newblock In \emph{Advances in Neural Information Processing Systems
  (NeurIPS)}, pages 7167--7177, 2018.

\bibitem[Liang et~al.(2018)Liang, Li, and Srikant]{liang2018enhancing}
Shiyu Liang, Yixuan Li, and R~Srikant.
\newblock Enhancing the reliability of out-of-distribution image detection in
  neural networks.
\newblock In \emph{International Conference on Learning Representations
  (ICLR)}, 2018.

\bibitem[Liu et~al.(2020)Liu, Wang, Owens, and Li]{liu2020energy}
Weitang Liu, Xiaoyun Wang, John Owens, and Yixuan Li.
\newblock Energy-based out-of-distribution detection.
\newblock In \emph{Advances in Neural Information Processing Systems
  (NeurIPS)}, volume~33, pages 21464--21475, 2020.

\bibitem[Madry et~al.(2018)Madry, Makelov, Schmidt, Tsipras, and
  Vladu]{madry2018towards}
Aleksander Madry, Aleksandar Makelov, Ludwig Schmidt, Dimitris Tsipras, and
  Adrian Vladu.
\newblock Towards deep learning models resistant to adversarial attacks.
\newblock In \emph{International Conference on Learning Representations
  (ICLR)}, 2018.

\bibitem[Nguyen et~al.(2015)Nguyen, Yosinski, and Clune]{nguyen2015deep}
Anh Nguyen, Jason Yosinski, and Jeff Clune.
\newblock Deep neural networks are easily fooled: High confidence predictions
  for unrecognizable images.
\newblock In \emph{Proceedings of the IEEE Conference on Computer Vision and
  Pattern Recognition}, pages 427--436, 2015.

\bibitem[Olah et~al.(2018)Olah, Satyanarayan, Johnson, Carter, Schubert, Ye,
  and Mordvintsev]{olah2018building}
Chris Olah, Arvind Satyanarayan, Ian Johnson, Shan Carter, Ludwig Schubert,
  Katherine Ye, and Alexander Mordvintsev.
\newblock The building blocks of interpretability.
\newblock \emph{Distill}, 3\penalty0 (3):\penalty0 e10, 2018.

\bibitem[Raghu et~al.(2017)Raghu, Gilmer, Yosinski, and
  Sohl-Dickstein]{raghu2017svcca}
Maithra Raghu, Justin Gilmer, Jason Yosinski, and Jascha Sohl-Dickstein.
\newblock {SVCCA}: Singular vector canonical correlation analysis for deep
  learning dynamics and interpretability.
\newblock In \emph{Advances in Neural Information Processing Systems
  (NeurIPS)}, pages 6076--6085, 2017.

\end{thebibliography}

\appendix

\section{Implementation Details}\label{apd:implementation}

\subsection{Network Architecture}

The monitor network consists of three main components: (1) Signal Encoder: processes probe activations to extract 112-dimensional features per probe; (2) Bidirectional GRU: 64 hidden units per direction with 0.3 dropout; (3) MLP Classifier: two layers (128→64→1) with ReLU activations and 0.5 dropout.

\subsection{Training Hyperparameters}

Batch size: 256; Learning rate: $10^{-3}$ with cosine decay; Weight decay: $10^{-4}$; Early stopping patience: 10 epochs; Gradient clipping: max norm 1.0.

\subsection{Computational Requirements}

Training time: approximately 8 hours on Apple M1 GPU for 40 epochs. Inference time: 15ms per image (signal encoder: 13ms, monitor: 2ms). Memory overhead: 5.2MB for monitor parameters.

\end{document}